# On pattern classification with weighted dimensions

Ayatullah Faruk Mollah[†]

*Department of Computer Science and Engineering, Aliah University*
*IIA/27 Newtown, Kolkata 700160, India*

**Abstract**. Studies on various facets of pattern classification is often imperative while working with multi-dimensional samples pertaining to diverse application scenarios. In this notion, weighted dimension-based distance measure has been one of the vital considerations in pattern analysis as it reflects the degree of similarity between samples. Though it is often presumed to be settled with the pervasive use of Euclidean distance, plethora of issues often surface. In this paper, we present (a) a detail analysis on the impact of distance measure norms and weights of dimensions along with visualization, (b) a novel weighting scheme for each dimension, (c) incorporation of this dimensional weighting schema into a KNN classifier, and (d) pattern classification on a variety of synthetic as well as realistic datasets with the developed model. It has performed well across diverse experiments in comparison to the traditional KNN under the same experimental setups. Specifically, for gene expression datasets, it yields significant and consistent gain in classification accuracy (around 10%) in all cross-validation experiments with different values of k. As such datasets contain limited number of samples of high dimensions, meaningful selection of nearest neighbours is desirable, and this requirement is reasonably met by regulating the shape and size of the region enclosing the k number of reference samples with the developed weighting schema and appropriate norm. It, therefore, stands as an important generalization of KNN classifier powered by weighted Minkowski distance with the present weighting schema.

## 1. Introduction

Since inception distance measure has been a vital consideration in pattern analysis as the distance between samples reflects the degree of their similarity. Though, this consideration is often presumed to be settled with the pervasive use of Euclidean distance, plethora of issues have been faced by researchers while working with multi-dimensional samples pertaining to various application scenarios. Consequently, appropriate twists and modifications have been made to come up with more generalized way of measurement. More importantly, manifold nature of application scenarios in general and pattern classification in specific has made it imperative to go deep and ponder over certain aspects like significance of individual dimension in representing inter-class and intra-class samples, their distributions, class imbalance, overlapped classes, cross-dimensional effect, etc.

Over the last couple of decades, some theoretical as well as empirical studies have been made upon the cardinal aspects mentioned above. While some studies have attempted to apply different weighted distance coupled with pattern classifiers, some attempt to combine benefit of multiple distance metrics. Geler et al. [1] demonstrated that weighted k-nearest neighbors (WKNN) approach is able to consistently outperform. They also presented the criteria of choosing neighborhood size k, constraint width parameter r, and a weighting scheme. Peng et al. [2] also proposed an improved WKNN algorithm, which is shown to outperform traditional algorithms including the K-nearest neighbor (KNN). Likewise, Rastin et al. [3] reported a weighting scheme called Prototype Weighting, which is applied on KNN and found to perform better particularly in case of imbalanced and overlapped classes. Hasan et al. [4] utilizes importance of dimensions with an ensemble approach for predicting the class of a recall sample.

---

[†] Corresponding author's email address: afmollah@aliah.ac.in

Among other studies, weighted Euclidean distance is applied for an anisotropic diffusion algorithm [5], multi-attributed graph for graph classification and clustering [6], plant layout selection [7], etc. A novel weighted Euclidean distance-based approach is reported by Rao et al. [8]. Bei et al. [9] attempted to combine the advantages of Euclidean distance, DTW distance and SPDTW distance. Apart from these, relevance of distance measure and cross-dimensional issues is reflected in deep learning and fuzzy space as well. Kalantidis et al. [10] have applied cross-dimensional weighting and aggregation of deep convolutional neural network layer outputs to create powerful image representation. Xiao [11] reported a distance measure for intuitionistic fuzzy sets and its application to pattern classification problems. Yang et al. [12] proposed a complex intuitionistic fuzzy ordered weighted distance measure.

In short, reasonable attention is already drawn towards distance measure, weighting scheme, cross-dimensional consideration, generalization of decision boundaries in imbalanced or overlapped distribution etc. However, specific focus on individual dimension of a multidimensional feature space is not evident so far. In this paper, we present (a) a detail analysis of norms of distance measure vis-a-vis dimensional weight along with visualization, (b) a novel weighting schema for each dimension, (c) application of this weighting schema into Minkowski distance equipped KNN classifier, and (d) pattern classification on a variety of synthetic and realistic datasets with the developed classification model.

## 2. Materials and Method

For a pattern classification problem, let $x = [x_1\ x_2\ ...\ x_n]^T$ and $y = [y_1\ y_2\ ...\ y_n]^T$ be two samples in $n$ dimensional space such that $x, y \in \mathbb{R}^n$, $\mathbb{R}$ is the set of real numbers. Then, distance between $x$ and $y$, denoted as $d(x, y)$ is obtained using various distance metrics. Euclidean distance, defined in Eq. 1, is the most popular metric used in almost all fields. It measures the length of the straight-line segment joining the two points $x$ and $y$.

$$d_{\text{euclidean}}(x, y) = \sqrt{\sum_{i=1}^{n}(x_i - y_i)^2} \qquad (1)$$

The city block distance, also called as Manhattan distance, and Chebyshev distance, also known as chess board distance are shown in Eq. 2 and Eq. 3 respectively.

$$d_{\text{manhattan}}(x, y) = \sum_{i=1}^{n}|x_i - y_i| \qquad (2)$$

$$d_{\text{chebyshev}}(x, y) = \max_{i}|x_i - y_i| \qquad (3)$$

Minkowski's distance between $x$ and $y$, as defined in Eq. 4, is a generalization of multiple distance metrics such as city block distance, Euclidean distance, chess board distance, etc.

$$d_{\text{minkowski}}(x, y) = \left(\sum_{i=1}^{n}|x_i - y_i|^p\right)^{1/p} \qquad (4)$$

It is also called as $p^{\text{th}}$ norm and denoted as $L_\text{p}$ metric or $L_\text{p}$ norm. It may be noted that when $p=1$, it refers to the city block distance, when $p = 2$, it refers to the Euclidean distance. And when $p \rightarrow \infty$, it represents the Chebyshev distance. So, Chebyshev distance may alternatively be expressed with Minkowski's distance as shown in Eq. 5.

$$d_{\text{chebyshev}}(x, y) = \lim_{p \rightarrow \infty}\left(\sum_{i=1}^{n}|x_i - y_i|^p\right)^{1/p} \qquad (5)$$

So, there are practically infinite number of distance measures. Understanding the geometric significance of these metrics is imperative in developing insights into design of pattern classification models. Fig. 1 demonstrates the impact of different norms on the obtained distance values by marking points having the same distance value, herein referred as *unidistat points*.

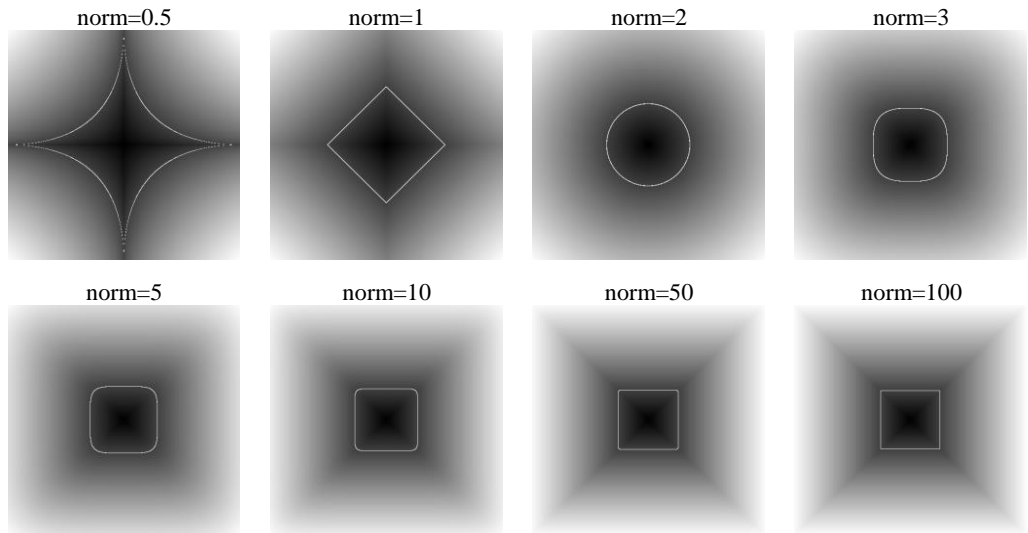

**Fig. 1** Points at equal Minkowski distance for different norms (Norm 1 signifies city block/ Manhattan distance, norm 2 signifies Euclidean distance, and with increase in norms the distances approach to $D_8$ /chessboard distance)

*2.1 Weighted Minkowski Distance*

Given, two points $x$ and $y$ in $\mathbb{R}^n$ where $w_i$ denotes the importance of $i^{\text{th}}$ dimension in some consideration, distance between them, i.e. $d(x,y)$ may be defined as

$$d(x,y) = \left(\sum_{i=1}^{n} w_i |x_i - y_i|^p\right)^{1/p} \quad (6)$$

where $\sum w_i = n$ and $w_i \geq 0$ for all $i$. It may be noted that it is a further generalization of Minkowski distance. When $w_i = 1$ for all $i$, it is equivalent to the ordinary Minkowski distance. Moreover, when $p = 1$, it is the Manhattan distance, when $p = 2$, it is the Euclidean distance, and when $p \to \infty$, it is the Chebyshev distance.

*Observation 1. Effect of dimensional weights are higher at lower norms*

When the value of $p$ is low, the dimensional weights $w_i$ play significant role in computing the distance. However, with the increase of the power parameter $p$, the value of the term $|x_i - y_i|^p$ increases rapidly, which normalizes the role of $w_i$. This may be clearly realized from Fig. 2.

*Observation 2. The limiting case*

With the increase in $p$, this metric approaches to the Chessboard distance, and in the limiting case i.e. when $p \to \infty$, $d(x,y) = d_{\text{chebyshev}}(x,y)$. Please note from Fig. 2 that when $p = 10$, computed distances are slightly varying with the change of dimension weights, and when $p = 100$, the distances are nearly equal in spite of changes in dimension weights.

*Observation 3. Shape of hypervolume*

In 3D, the enclosed area is a volume elongated towards axes of lesser weights. Likewise, when number of dimensions is more than 3, *unidistant points* form a hypervolume with proportionate elongation.

*2.2 Use-case Scenario*

Computation of distance is an important consideration in many fields including pattern recognition and machine learning. There may be situations when all dimensions representing certain entities may not be equally important. In certain situations, some dimensions may be redundant as well. It can be intuitively realized that in those situations, computation of distance between two points should not treat all dimensions equally. Distance along dimensions having greater importance should contribute more in the computed distance value. Below, we mention a practical situation where this consideration is relevant.

Given, two or more distributions representing unique category in $n$ dimensional space, all dimensions may not be equally discriminatory. Let, $w_i$ denote the importance of $i^{\text{th}}$ dimension in discriminating the distributions. Then, the proposed metric may be helpful, which may be realized from Fig. 3.

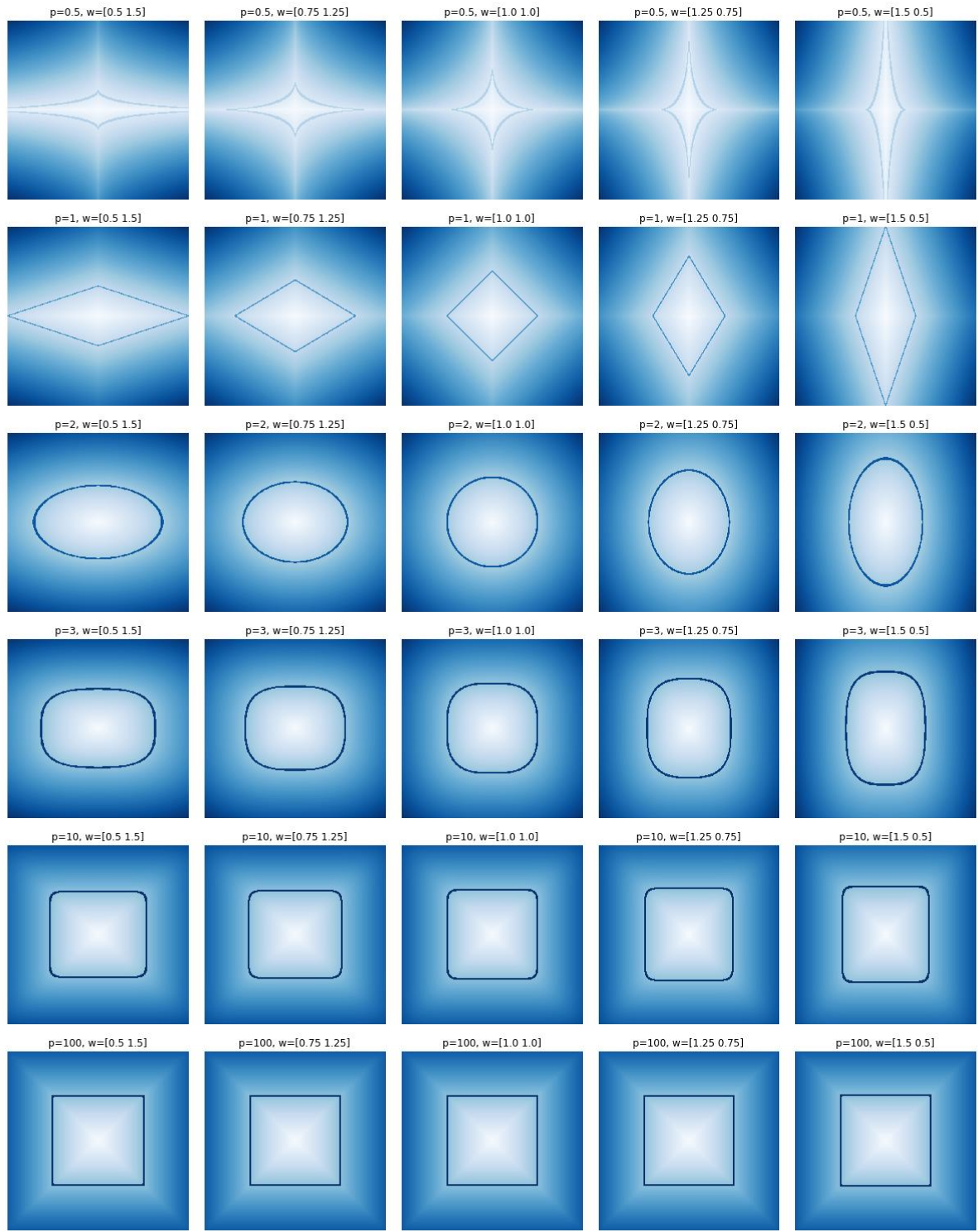

**Fig. 2** Computed distances at various locations with varying norm and dimension weights. Neighborhoods with respect to the origin have been marked by boundary constituted with unidistant points. Effect of dimensional weights are higher at lower norms.

## 2.3 Setting Weights to Dimensions

A weighting schema is therefore affable to stimulate the dimensions of higher importance and weaken the dimensions of lower importance. A novel weighting scheme shown in Eq. 7 is introduced for this purpose.

$$w_i = \kappa + (1 - \kappa) * n * \frac{\lambda_i}{\sum \lambda_i} \quad (7)$$

where $\kappa$ ('kappa') is a real constant in the range [0.0,1.0] and $\lambda_i$ denotes the weight or importance of $i^{th}$ dimension. Computing importance of dimensions is an important issue and it depends on the domain of application. In the context of pattern analysis, values of $\lambda_i$ could be set by following any univariate feature fitness function such as ANOVA f-statistic, mutual information, symmetrical uncertainty, Fisher score, Gini index, etc. In this study, the fitness function adopted for multivariate and multiclass scenario is as shown in Eq. 8. It may be noted that in this weighting schema, adequate flexibility may be exercised through $\kappa$.

$$\lambda_i = \sum_{p(s,t)} \frac{|\mu(\omega_s) - \mu(\omega_t)|}{\sigma(\omega_s) + \sigma(\omega_t)} \quad : p(s,t) \text{ is a unique pair in } n \times n \quad (8)$$

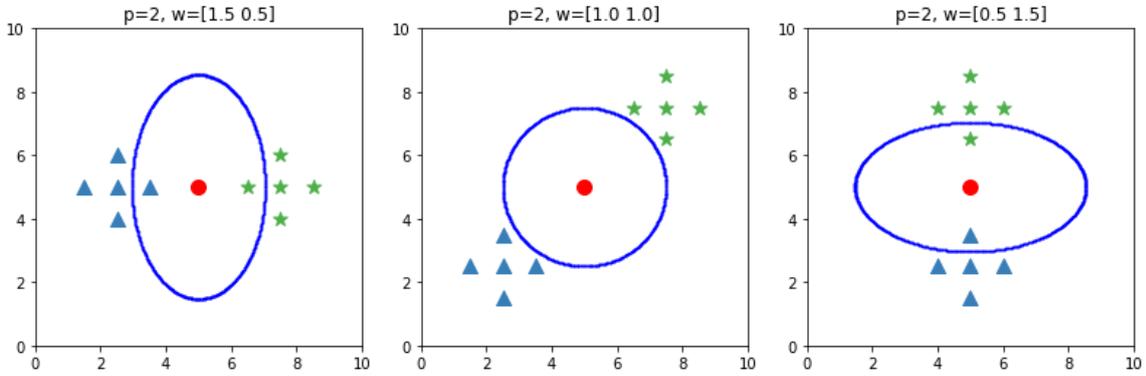

**Fig. 3** Illustration on the impact of dimensional weights while the norm is constant (It may be noted that when weights are same it is the case of Euclidean distance, but for other weight the region varies depending on the weight values).

## 2.4 Illustration on Synthetic Dataset

In a pattern classification scenario, all dimensions do not characterize the samples at equal scale. The dimension along which samples are scattered or distributed the most contributes the most in discriminating the samples of different classes. On the contrary, the dimension along which samples of different classes are overlapping poorly contributes, rather misleads in discriminating samples of between-classes. The proposed schema assigns weightage to dimensions in due proportion of their discriminating abilities.

The schema also provides a tuning parameter $\kappa$ which can be used to rely on the weighting schema at different degree. When $\kappa = 0$, it signifies that we fully rely on the weighting scheme. On the other side, when $\kappa = 1$, weights of all dimensions turn to 1.0 which signifies that we do not rely on the weighting scheme at all. Thus, it stands as a fascinating generalization of dimensional weights. In Fig. 4, a two-class synthetic dataset is considered to show the scatter of data along the two dimensions. It may be observed that the volume containing the reference samples significantly varies with the norm and dimensional weights. For an unknown sample marked as a red dot, appropriate reference samples are found along the x-direction since samples are largely scattered along this axis.

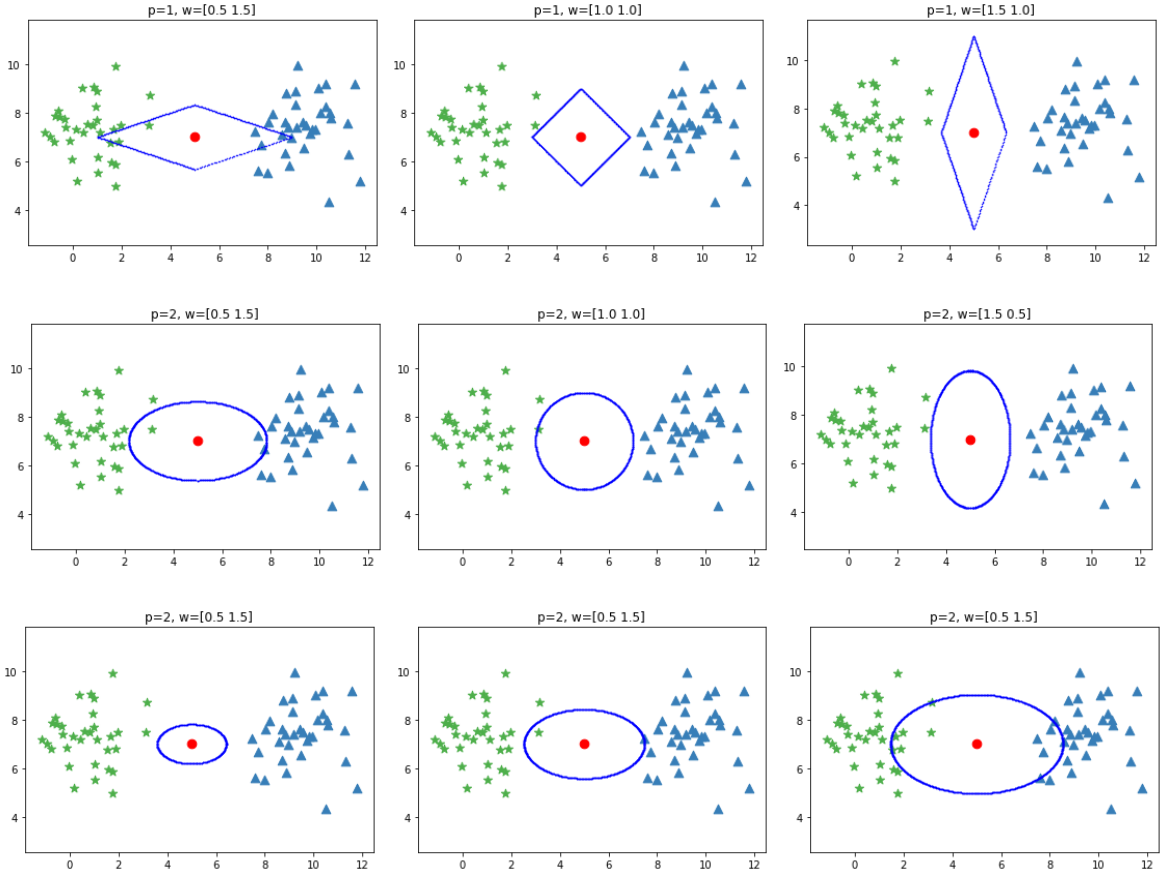

**Fig. 4** Impact of the dimensional weights and norm on the shape of the hyper-volume containing the k number of reference samples in a KNN classifier (The dimension along which samples are more scattered contributes more in finding appropriate samples for majority voting).

## 3. Experiments and Observations

Experiments have been conducted on five publicly available real-world datasets viz. iris [13], breast cancer [13], diabetes [14], leukemia [15] and colon cancer [16], by applying the developed dimensional weighting schema on a k-nearest neighbors classifier. It may be noted that for all these datasets, experiments have been carried out for multiple values of k and multiple folds of the datasets. The same set of experiments have been conducted with the traditional KNN as well for comparison.

### 3.1 Classification Performance

Classification performance of the traditional KNN vis-a-vis the developed weighting scheme equipped KNN for all the five datasets is presented in Table 1-5. Both sets of experiments have been carried out for odd number of neighbors i.e. k=1,3,5 for breaking tie situation, if arises in majority voting, following three different stratified cross-validation approaches (3-folds, 5-folds and 10-folds). Moreover, in all these experiments, $\kappa = 0$ which indicates that weight assignment scheme is fully utilized to leverage its true potential.

From these experimental results, it is evident that the developed scheme performs significantly better in two datasets – leukemia and colon cancer. More interestingly, in all the 9 experiments (3 different folds for 3 different k values) on these datasets, it has yielded better classification accuracy reflecting its consistence in superior performance. In the other datasets namely iris, breast cancer and diabetes, the developed method performed slightly better. It is interesting to note that, in no case, the developed method performed poor than the traditional KNN.

**Table 1**. Classification accuracy of the traditional KNN and the proposed weighted KNN on Iris dataset [13] for 3-folds, 5-folds and 10-folds cross validation experiments with different values of k.

| k | n-fold | Accuracy | |
|---|---|---|---|
| | | KNN (Euclidean) | KNN (Proposed Distance) |
| 1 | 3 | 0.9467 | 0.9667 |
| | 5 | 0.9467 | 0.9667 |
| | 10 | 0.9533 | 0.9600 |
| 3 | 3 | 0.9400 | 0.9667 |
| | 5 | 0.9533 | 0.9600 |
| | 10 | 0.9533 | 0.9600 |
| 5 | 3 | 0.9600 | 0.9667 |
| | 5 | 0.9600 | 0.9667 |
| | 10 | 0.9533 | 0.9600 |
| Mean | | **0.9518** | **0.9637** |
| sd | | **0.006098472** | **0.003329257** |

**Table 2**. Average classification accuracy of the proposed weighted KNN vis-a-vis traditional KNN on breast cancer dataset [13].

| k | n-fold | Accuracy | |
|---|---|---|---|
| | | KNN (Euclidean) | KNN (Proposed Distance) |
| 1 | 3 | 0.9472 | 0.9526 |
| | 5 | 0.9508 | 0.9561 |
| | 10 | 0.9508 | 0.9579 |
| 3 | 3 | 0.9596 | 0.9631 |
| | 5 | 0.9578 | 0.9613 |
| | 10 | 0.9648 | 0.9631 |
| 5 | 3 | 0.9613 | 0.9613 |
| | 5 | 0.9649 | 0.9666 |
| | 10 | 0.9666 | 0.9666 |
| Mean | | **0.9582** | **0.9610** |
| sd | | **0.006677158** | **0.004420687** |

**Table 3**. Pima Indians Diabetes [14] prediction with the developed weighting scheme equipped with KNN classifier.

| k | n-fold | Accuracy | |
|---|---|---|---|
| | | KNN (Euclidean) | KNN (Proposed Distance) |
| 1 | 3 | 0.7057 | 0.7188 |
| | 5 | 0.7123 | 0.6967 |
| | 10 | 0.7057 | 0.7096 |
| 3 | 3 | 0.7383 | 0.7344 |
| | 5 | 0.7423 | 0.7201 |
| | 10 | 0.7434 | 0.7460 |
| 5 | 3 | 0.7318 | 0.7409 |
| | 5 | 0.7331 | 0.7423 |
| | 10 | 0.7382 | 0.7565 |
| Mean | | **0.7279** | **0.7295** |
| sd | | **0.014655829** | **0.018238407** |

**Table 4**. Classification performance of the developed weighting and classification approach on the publicly available 2-class leukemia dataset [15] (Significant improvement over traditional KNN may be noted).

| k | n-fold | Accuracy | |
|---|---|---|---|
| | | KNN (Euclidean) | KNN (Proposed Distance) |
| 1 | 3 | 0.8194 | 0.9444 |
| | 5 | 0.8038 | 0.9010 |
| | 10 | 0.8589 | 0.9571 |
| 3 | 3 | 0.7778 | 0.9444 |

|   |     | 5  | 0.8305 | 0.9429 |
|---|-----|----|--------|--------|
|   |     | 10 | 0.8446 | 0.9571 |
| 5 |     | 3  | 0.7917 | 0.9306 |
|   |     | 5  | 0.7895 | 0.9295 |
|   |     | 10 | 0.8054 | 0.9589 |
|   | Mean |   | **0.8135** | **0.9407** |
|   | sd   |   | **0.025468082** | **0.017336225** |

**Table 5**. Colon cancer [16] classification with the developed method for 3 different cross-validation experiments with 3 different values of k (Substantial improvement may be noted in all experiments).

| k | n-fold | Accuracy | |
|---|--------|----------|---|
|   |        | KNN (Euclidean) | KNN (Proposed Distance) |
| 1 | 3  | 0.6913 | 0.8056 |
|   | 5  | 0.8077 | 0.8859 |
|   | 10 | 0.7190 | 0.7857 |
| 3 | 3  | 0.6786 | 0.8056 |
|   | 5  | 0.8397 | 0.8705 |
|   | 10 | 0.7690 | 0.8214 |
| 5 | 3  | 0.7103 | 0.8214 |
|   | 5  | 0.7910 | 0.8705 |
|   | 10 | 0.7381 | 0.8381 |
| Mean |  | **0.7494** | **0.8339** |
| sd   |  | **0.052320178** | **0.032724884** |

## 3.2 Discussion

A summary of performance in terms of the mean and standard deviation of classification accuracies of all experiments on the five datasets considered in this work is shown in Fig. 5. It is clearly evident that performance of the developed method is slightly better in iris, breast cancer and diabetes datasets, while that of the leukemia and color cancer is considerably higher.

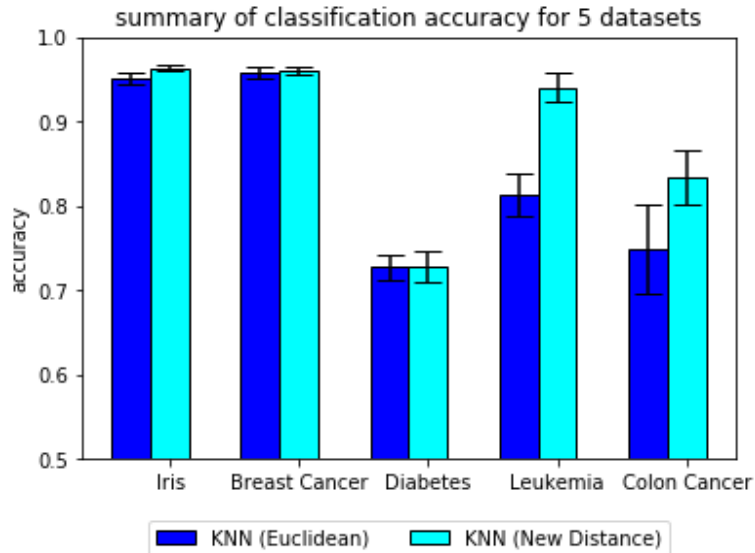

**Fig. 5** Summary of classification performance of traditional KNN vis-a-vis developed weighting scheme equipped KNN across five datasets (mean accuracy and standard deviation have been shown).

A closer observation revealed that unlike the first three real-world datasets, leukemia and colon cancer datasets are of a different genre (i.e. gene expression data). As gene expression datasets usually have high number of dimensions and low number of samples, the underlying considerations of the developed method has appeared to be very relevant. In such situations, weakening of less important features and strengthening more important features is really needed. Moreover, selection of k number of reference samples around an unknown test sample, which is a key factor in predicting the class of the test sample, is better realized with the developed approach.

## 4. Conclusion

In this paper, (i) a study on the impact of distance measure norms and weights of dimensions is made through visualization as well as experimentation, (ii) a novel weighting schema is presented for determining weight in proportion to their relevance in discriminating samples of between-classes in feature space, (iii) this weighting schema is incorporated into a KNN classifier, and (iv) evident benefits on five real-world datasets have been achieved. Specifically, on gene expression datasets, this method yields significant gain in classification accuracy (around 10%), and to our surprise, such gain is found consistent in all cross-validation experiments with different values of k of a KNN classifier. Further investigation revealed that such datasets contain limited number of samples of very high dimensions, in which meaningful selection of k number of nearest neighbors is imperative. In the developed method, one can regulate the shape and size of the region to enclose the k number of reference samples by tuning norms and weighting constant 'kappa', while at the bottom line it may be made to act as a traditional KNN. Thus, it stands as an important generalization of KNN classifier powered by weighted Minkowski distance with the presented novel weighting schema.